\documentclass[lettersize,journal]{IEEEtran}
\usepackage{amsmath,amsfonts}
\usepackage{algorithmic}
\usepackage{algorithm}
\usepackage{array}
\usepackage[caption=false,font=normalsize,labelfont=sf,textfont=sf]{subfig}
\usepackage{textcomp}
\usepackage{stfloats}
\usepackage{url}
\usepackage{verbatim}
\usepackage{graphicx}
\usepackage{cite}

\usepackage{booktabs}
\usepackage{multirow}

\usepackage{hyperref}

\usepackage{xcolor}
\begin{document}
	
	\title{Occupancy-MAE: Self-supervised Pre-training Large-scale LiDAR Point Clouds with Masked Occupancy Autoencoders}
	
	\author{Chen~Min, Liang~Xiao, Dawei~Zhao, Yiming~Nie, Bin~Dai
		\thanks{This work was supported by the National Natural Science Foundation of China under Grant Nos. 61790565 and 61803380.}
		\thanks{C. Min is with the School of Computer Science, Peking University, Beijing 100071, China (e-mail: minchen@stu.pku.edu.cn).}
		\thanks{L. Xiao, D. Zhao, Y. Nie, and B. Dai are with the Unmanned Systems Technology Research Center, Defense Innovation Institute, Beijing 100071, China (e-mail: xiaoliang@nudt.edu.cn;  adamzdw@163.com; e\_ming\_e\_ming@163.com; ibindai@163.com).}}
	%
	\markboth{Journal of \LaTeX\ Class Files,~Vol.~14, No.~8, August~2021}%
	{Shell \MakeLowercase{\textit{et al.}}: A Sample Article Using IEEEtran.cls for IEEE Journals}
	
	\IEEEpubid{0000--0000/00\$00.00~\copyright~2021 IEEE}
	
	\maketitle
	
	\begin{abstract}
		Current perception models in autonomous driving heavily rely on large-scale labelled 3D data, which is both costly and time-consuming to annotate. This work proposes a solution to reduce the dependence on labelled 3D training data by leveraging pre-training on large-scale unlabeled outdoor LiDAR point clouds using masked autoencoders (MAE). While existing masked point autoencoding methods mainly focus on small-scale indoor point clouds or pillar-based large-scale outdoor LiDAR data, our approach introduces a new self-supervised masked occupancy pre-training method called Occupancy-MAE, specifically designed for voxel-based large-scale outdoor LiDAR point clouds.
		Occupancy-MAE takes advantage of the gradually sparse voxel occupancy structure of outdoor LiDAR point clouds and incorporates a range-aware random masking strategy and a pretext task of occupancy prediction. By randomly masking voxels based on their distance to the LiDAR and predicting the masked occupancy structure of the entire 3D surrounding scene, Occupancy-MAE encourages the extraction of high-level semantic information to reconstruct the masked voxel using only a small number of visible voxels.
		Extensive experiments demonstrate the effectiveness of Occupancy-MAE across several downstream tasks. For 3D object detection, Occupancy-MAE reduces the labelled data required for car detection on the KITTI dataset by half and improves small object detection by approximately 2\% in AP on the Waymo dataset. For 3D semantic segmentation, Occupancy-MAE outperforms training from scratch by around 2\% in mIoU. For multi-object tracking, Occupancy-MAE enhances training from scratch by approximately 1\% in terms of AMOTA and AMOTP.
		Codes are publicly available at \url{https://github.com/chaytonmin/Occupancy-MAE}.
		
	\end{abstract}
	
	
	\section{Introduction}
	
	\label{sec:intro}
	Accurate 3D perception is a core technique in autonomous driving, as it enables vehicles to obtain precise information about their surroundings~\cite{radar}. Numerous large-scale outdoor LiDAR point cloud datasets such as KITTI~\cite{kitti}, Waymo~\cite{waymo}, nuScenes~\cite{nuscenes}, and ONCE~\cite{once} have been published, showcasing the potential of environmental perception for unmanned vehicles. However, the collection and annotation of large-scale LiDAR point clouds for common tasks like 3D object detection and semantic segmentation can be extremely time-consuming and labour-intensive. For instance, skilled workers can only annotate around 100-200 frames per day~\cite{once}. Therefore, self-supervised learning using large-scale unannotated LiDAR point clouds is crucial for enhancing the perceptual ability of autonomous driving. It may pave the way for developing the next-generation, industry-level autonomous driving perception model that is both powerful and robust~\cite{once}.
	
	\begin{figure}[t]
		\centering
		\includegraphics[width=0.48\textwidth]{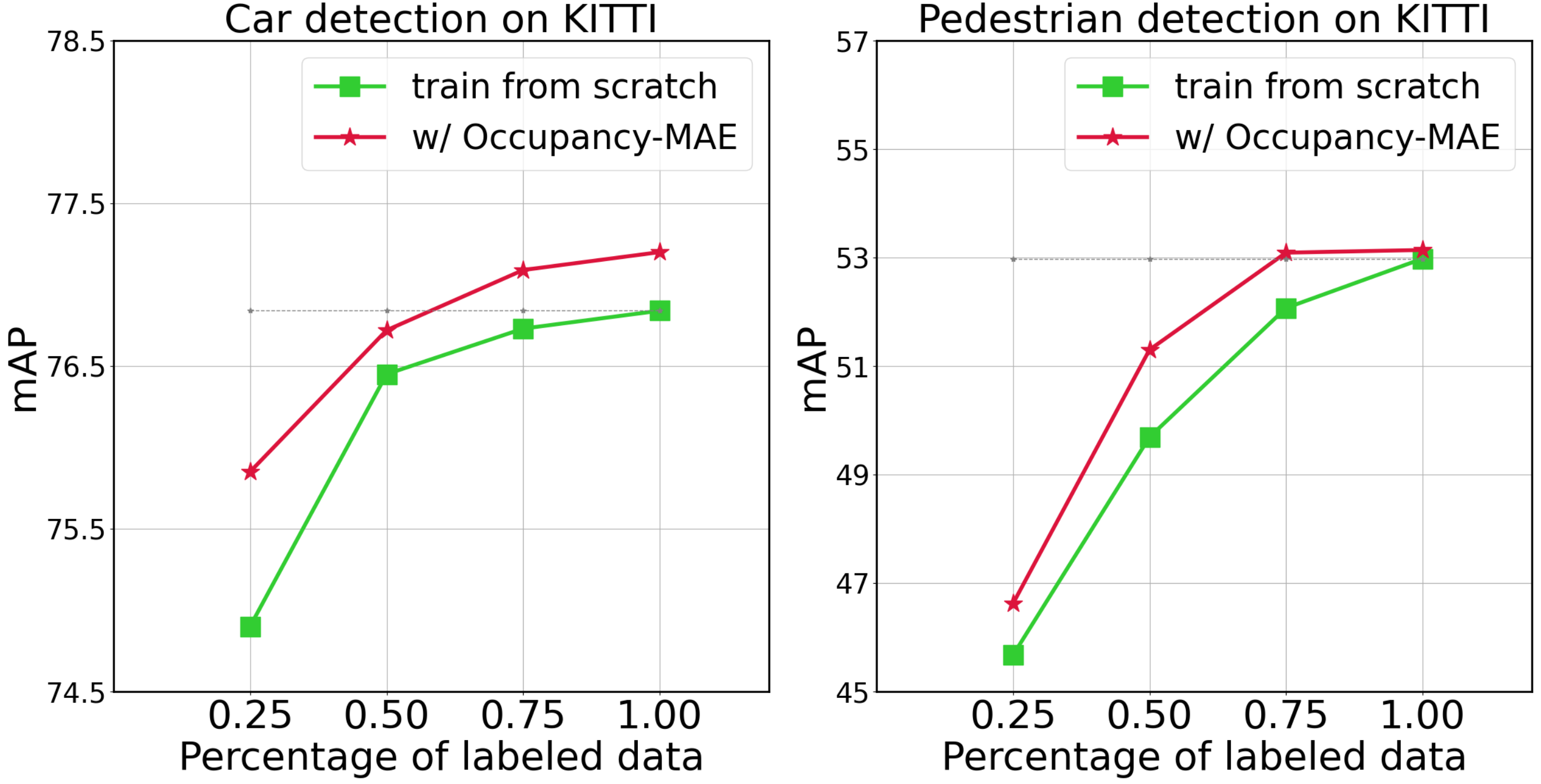} 
		\caption{
			$\textbf{Label-efficiency of our self-supervised pre-training}$. Occupancy-MAE outperforms training from scratch and achieves the same detection performance with fewer labelled data (about 50\% for the car class and 75\% for the pedestrian class).}
		\label{fig:percentage}
	\end{figure}
	
	\begin{figure*}[t]
		\centering
		\includegraphics[width=0.99\textwidth]{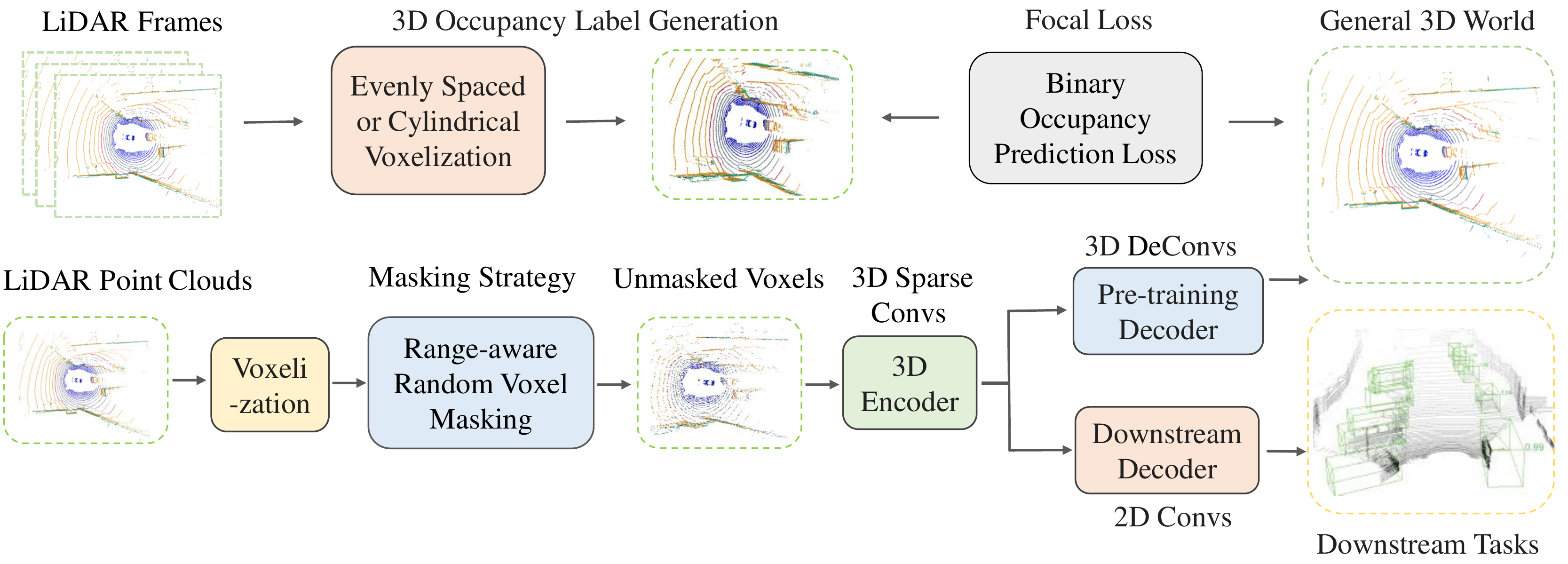}
		\caption{The overall architecture of our Occupancy-MAE. We first transform the large-scale irregular LiDAR point clouds into volumetric representations, randomly mask the voxels according to their distance from the LiDAR sensor (i.e., range-aware masking strategy), then reconstruct the geometric occupancy structure of the general 3D world with an asymmetric autoencoder network. We adopt the 3D Spatially Sparse Convolutions~\cite{second} with positional encoding as the encoding backbone. We apply binary occupancy classification as the pretext task to distinguish whether the voxel contains points. After pre-training, the lightweight decoder is discarded, and the encoder is used to warm up the backbones of downstream tasks.
		}
		\label{fig:flowchart}
	\end{figure*}
	\IEEEpubidadjcol
	
	Self-supervised learning has made significant strides in recent years, enabling the pre-training of rich features without human annotations. The simple masked autoencoding approach has been particularly effective in learning representative features, where the task is to reconstruct masked data from unmasked input~\cite{bert,mae,cae,simmim}. In natural language processing, masked autoencoding has enabled the training of large language models, such as BERT~\cite{bert}. Similarly, in 2D vision, masked autoencoding has outperformed supervised pre-training counterparts~\cite{mae}.
	
	Recently, several works on masked point autoencoding have been proposed, such as Point-MAE~\cite{point_mae}, Point-BERT~\cite{point_betr}, MaskPoint~\cite{maskpoint}, and Point-M2AE~\cite{pointm2ae}. However, these methods have primarily focused on small-scale point clouds, such as synthetic point cloud datasets like ShapeNet~\cite{shapenet} and indoor point cloud datasets like ScanNet~\cite{scannet}. In contrast, masked autoencoding for large-scale outdoor LiDAR point clouds~\cite{kitti, waymo, nuscenes, once}, has received less attention. Training from scratch on large-scale labelled LiDAR point clouds remains the dominant approach~\cite{once}. To introduce the idea of masked autoencoding to self-supervised learning on large-scale outdoor LiDAR point clouds, we first identify the challenges compared to masked small-scale point cloud autoencoding works~\cite{point_betr,point_mae,maskpoint,pointm2ae}, and then propose solutions:
	
	Firstly, small-scale point clouds~\cite{shapenet,scannet} differ from large-scale outdoor LiDAR point clouds~\cite{kitti,waymo,nuscenes, once} in several ways. (1) Small-scale point clouds typically contain far fewer points than their large-scale counterparts, with ShapeNet containing approximately 2k points and ScanNet downsampled to about 2k points for use in masked small-scale point cloud autoencoding works~\cite{point_betr,point_mae,maskpoint,pointm2ae}. In contrast, LiDAR sensors such as the Velodyne HDL-64E LiDAR can scan up to 192,000 points per frame, covering an area of 160$\times$160$\times$20 meters~\cite{squeezesegv3}. (2) Small-scale point clouds are often evenly distributed, while large-scale LiDAR point clouds become sparser as the distance from the LiDAR sensor increases. (3) Existing masked small-scale point cloud autoencoding works~\cite{point_betr,point_mae,maskpoint,pointm2ae} typically rely on furthest point sampling (FPS) and k-nearest neighbours (kNN) to divide points into equal subsets, which is not suitable for large-scale outdoor LiDAR point clouds. The challenge of processing such large-scale point clouds efficiently, and in real-time, has been highlighted in previous works~\cite{cylinder3d,squeezesegv3}, making pre-training on these datasets a more challenging task.
	
	Secondly, existing masked small-scale point clouds autoencoding works~\cite{point_betr,point_mae,pointm2ae} mainly focus on regressing the missing points. However, it is not appropriate to reconstruct the features of large-scale LiDAR point clouds through regression, as these features contain essential spatial information~\cite{second}. Moreover, the positional encoding of voxels can offer a shortcut for the decoder~\cite{pointconstrast}. To address this issue, we shift our focus to the occupancy distribution of large-scale LiDAR point clouds and design the occupancy prediction objective as the pretext task. By doing so, the network is forced to learn representative features to recover the overall structure of the 3D scene, making this simple task an effective solution for our pre-training method.
	
	Thirdly, the random masking strategy used in existing masked small-scale point clouds autoencoding works~\cite{point_betr,point_mae,maskpoint,pointm2ae} may not be suitable for large-scale outdoor LiDAR point clouds due to their uneven distribution. Unlike small-scale point clouds, the density of large-scale LiDAR point clouds varies based on the distance to the LiDAR sensor. Hence, a uniform random masking strategy for all voxels would not be optimal. We propose a range-aware random masking strategy for large-scale LiDAR point clouds to address this. This strategy adjusts the masking ratio based on the voxel's distance to the LiDAR sensor, with the masking ratio decreasing as the distance from the sensor increases.
	
	Transformer has started to make their presence felt in processing point clouds~\cite{li2023transformer}. For instance, PCT~\cite{pct} performs global self-attention across the entire point cloud, while PointASNL~\cite{yan2020pointasnl}, PointFormer~\cite{mao2021voxel}, and VoxSet~\cite{he2022voxel} apply transformer-based architectures to the local neighborhood of each point. However, the efficiency of these local transformers is constrained by neighborhood query and feature restructuring. SST~\cite{sst} and FlatFormer~\cite{liu2023flatformer} apply window-based point cloud transformers. These models project the point cloud into a bird's-eye view (BEV) and partition the BEV space into non-overlapping windows of the same spatial size. However, these approaches could lose valuable information in the vertical axis. We use 3D Sparse Convolutions~\cite{second} with a positional encoding module in the encoder. This allows the network to focus on visible 3D voxels instead of processing the entire point cloud, significantly reducing the computational cost. The positional encoding module, similar to the positional embeddings used in Transformers, encodes the spatial information of the voxels into a fixed-size embedding vector. 
	
	Recently, there have been several proposed self-supervised pre-training algorithms (e.g., Voxel-MAE~\cite{voxelmae}, MV-JAR~\cite{mv-jar}, GeoMAE~\cite{geomae}, and GD-MAE~\cite{gd-mae}) based on masked autoencoder for large-scale outdoor LiDAR point clouds. However, these algorithms are specifically tailored for pre-training pillar-based outdoor LiDAR point clouds method SST~\cite{sst} (SST replaces Pointpillars~\cite{pointpillars} backbone with Transformer~\cite{transformer}
	and cannot process 3D voxels.), overlooking the crucial height information of the point clouds. 
	
	Information redundancy in LiDAR point clouds refers to the presence of duplicate or highly similar data points within the point cloud dataset~\cite{sun2020novel,tu2019point}. This redundancy can arise from various sources~\cite{sun2019novel,feng2020real}, such as overlapping LiDAR swaths, multiple returns, sampling density, sensor artifacts, and noise. The overlapping points contain similar information about the terrain, which can be redundant. The multiple returns often contain information about the same physical feature from different angles, leading to redundancy. The high sampling density can lead to redundancy, as multiple points may describe the same ground surface, particularly in flat or uniform areas. The artifacts can result in redundant data points that do not represent actual features on the ground. Consequently, a substantial amount of redundant spatial information remains concealed within LiDAR point clouds, making them suitable for masked autoencoding methods to learn rich and representative features.
	
	Occupancy as a compressed representation of LiDAR point clouds can be defined as whether or not a voxel in the 3D grid contains points. Occupancy prediction is a commonly used approach in general 3D world representation for autonomous driving tasks~\cite{occupancy}. It involves dividing the environment into a grid and estimating the probability of each cell being occupied or free. This approach has been applied in various tasks such as obstacle detection, path planning, and simultaneous localization and mapping (SLAM)~\cite{fu2020fast,occupancyflow,hoermann2018dynamic}. 
	
	\IEEEpubidadjcol
	
	Driven by the above analyses, we propose the first self-supervised masked occupancy autoencoding framework, called $\textbf{Occupancy-MAE}$, for pre-training large-scale outdoor LiDAR point clouds. Figure~\ref{fig:flowchart} illustrates the workflow of our Occupancy-MAE, which first employs the range-aware masking strategy to mask voxels randomly, and then feeds the unmasked voxels into the 3D sparse encoder. The output of the 3D decoder is the probability that each voxel contains points, and we calculate the binary occupancy classification loss to pre-train the network. Training on the pretext task of masked occupancy classification encourages the encoder network to be voxel-aware of the entire object shape, thereby learning representative features for 3D perception.
	
	After analyzing the results of our experiments, we conclude that Occupancy-MAE is a simple and effective self-supervised learning framework that can generalize well across various downstream tasks. In the 3D object detection task, our method outperforms the state-of-the-art self-supervised learning methods by 0.5\% $\sim$ 6\% mAP on the ONCE dataset, which is a recently published dataset for self-supervised learning on large-scale LiDAR point clouds. Moreover, pre-training with Occupancy-MAE can significantly improve the performance of popular 3D detectors such as SECOND~\cite{second}, PV-RCNN~\cite{pv_rcnn}, CenterPoint~\cite{centerpoint}, and PV-RCNN++~\cite{pv_rcnn++} trained from scratch on KITTI, Waymo, and nuScenes datasets, especially for small objects. As illustrated in Figure~\ref{fig:percentage}, Occupancy-MAE is a data-efficient learner that can effectively train large-scale LiDAR point clouds with limited annotated 3D data. For the 3D semantic segmentation task, Occupancy-MAE with a two-layer decoder can improve the training from scratch by about 2\% mIoU. For multi-object tracking task, Occupancy-MAE enhances training from scratch by approximately 1\% in terms of AMOTA and AMOTP. We also demonstrate the effectiveness of our approach in the unsupervised domain adaptation task, which confirms the transfer learning ability of Occupancy-MAE. Even with a 90\% masking ratio, Occupancy-MAE can still learn representative features as large-scale LiDAR point clouds are information redundant, which ultimately improves the performance of 3D perception.
	
	The main contributions of this work are listed below:
	\begin{itemize}
		\item We present a novel self-supervised masked occupancy autoencoding framework called Occupancy-MAE, for pre-training large-scale outdoor LiDAR point clouds, reducing the need for expensively annotated 3D data.
		\item We propose a 3D occupancy prediction pretext task that leverages the gradually sparse occupancy structure of large-scale LiDAR point clouds. By recovering the masked occupancy distribution of the 3D surrounding world from a small number of visible voxels, the network is forced to extract high-level semantic information. 
		\item We introduce the range-aware random masking strategy to take advantage of the varying density of large-scale LiDAR point clouds, which improves the pre-training performance.
		\item Our proposed Occupancy-MAE significantly outperforms training from scratch on various downstream tasks, including 3D object detection, semantic segmentation, multi-object tracking, and unsupervised domain adaptation.
	\end{itemize}
	\section{Related Work}
	\subsection{LiDAR-based 3D Perception}
	LiDAR-based 3D perception models with accurate 3D spatial information have been widely used in the field of autonomous driving~\cite{uni3d,adpt}. LiDAR-based 3D object detectors can be categorized into point-based~\cite{pointrcnn, 3dssd}, voxel-based~\cite{voxelnet, second, pointpillars}, and point-voxel-based~\cite{pv_rcnn, fast_point_rcnn}. Point-based object detectors extract discriminative features from raw point clouds with PointNet~\cite{pointnet} and generate proposals centred at each point with high computation cost. Voxel-based detectors transform the irregular point clouds into volumetric representations, which will degrade the fine-grained localization accuracy. Point-voxel-based methods take advantage of the localization accuracy of point-based detectors and the computational efficiency of voxel-based detectors. LiDAR-based 3D segmentation methods are divided into grid-based~\cite{squeezeseg, rangenet} and voxel-based~\cite{cylinder3d, spvnas}. Grid-based methods focus on converting the 3D point clouds into a 2D frontal-view image or a range image, which fails to model the 3D geometric information. Voxel-based methods convert the point clouds into volumetric representations. The existing 3D perception methods are trained with large-scale labelled 3D data. How to design the self-supervised learning network to minimize dependence on 3D annotation has rarely been studied.
	
	\subsection{Self-supervised Learning}
	Self-supervised Learning (SSL) has gained popularity in recent years as an effective approach to avoiding costly data annotation. The pretext task of predicting the relative location of image patches was proposed in \cite{patch_id}. Methods in \cite{rotation, xu1, xu2} have designed rotation prediction tasks, which have shown promising results in learning representative features. In \cite{jigsaw}, a jigsaw puzzle prediction task is introduced, which generalizes well in domain adaptation for object recognition. DeepCluster~\cite{deepcluster} and SwAV~\cite{swav} obtain pseudo labels with k-means clustering and use these labels to train networks. Other methods such as Moco~\cite{moco}, PointContrast~\cite{pointconstrast}, BYOL~\cite{byol}, ProposalContrast~\cite{proposalcontrast}, and DepthContrast~\cite{depthconstrast} construct contrastive views for self-supervised learning. Recently, MAE~\cite{mae} has shown promising results in self-supervised learning by first masking random patches of the input image and then reconstructing the missing pixels with a simple autoencoder framework. VideoMAE~\cite{videomae} extends the MAE into spatiotemporal representation learning from videos, which have greater information redundancy. ALSO~\cite{also} trains the model on a pretext task which is the reconstruction of the surface on which the 3D points are sampled. ISCC~\cite{implicit} applies contrastive cluster and implicit surface reconstruction for self-supervised pre-training large-scale outdoor LiDAR point clouds. Our Occupancy-MAE follows the design philosophy of MAE and applies it to large-scale outdoor LiDAR point clouds based on their geometric characteristics, such as sparsity and varying density.
	
	\subsection{Masked autoencoders for point clouds}
	Masked autoencoding has achieved success in NLP~\cite{bert} and image~\cite{mae}, leading to the development of masked autoencoding techniques for point clouds in the last year. Point-BERT~\cite{point_betr} first introduces MAE to pre-train small-scale point clouds. Point-MAE~\cite{point_mae} reconstructs the small-scale point patches with the Chamfer distance. MaskPoint~\cite{maskpoint} designs the decoder for discriminating the small-scale masked point patches. Point-M2AE~\cite{pointm2ae} applies the pyramid architectures to model spatial geometries and capture both fine-grained and high-level semantics of 3D shapes. However, these methods focusing on small-scale indoor point clouds cannot process large-scale outdoor LiDAR point clouds under the property of a large range of scenes and varying density. Voxel-MAE~\cite{voxelmae}, MV-JAR~\cite{mv-jar}, GeoMAE~\cite{geomae}, and GD-MAE~\cite{gd-mae} adopt masked autoencoding for large-scale point clouds but are limited to only pillar-based methods, which will lose vertical information. Our proposed Occupancy-MAE overcomes these limitations and enables the pre-training of large-scale LiDAR point clouds for both voxel and pillar-based methods.
	\section{Methodology}
	
	\begin{table*}[t]
		\centering
		\caption{The details of our Occupancy-MAE architecture which consists of a 3D Encoder and a 3D Decoder. 3DSparseCvonv and 3DTransCvonv denote the 3D Sparse Convolutional layer proposed in SECOND [45] and the common 3D Deconvolutional layer, respectively. Here we display the output size for pre-training SECOND on the KITTI dataset.}
		{
			\begin{tabular}{c|c|c|c}
				\toprule
				\textbf{Input} &\textbf{Layer} \textbf{Destription} &\textbf{Output} &\textbf{Output Size} \\
				\midrule
				\midrule
				Input point clouds &Mean operator & Voxel representation & 1600$\times$ 1408 $\times$ 41 $\times$ 4 \\
				\midrule
				\midrule
				\multicolumn{4}{c} {3D Encoder} \\
				\midrule
				\midrule
				Masked voxels $ \textbf{V}_{input} $ & 3DSparseCvonv, filter=(3,3,3), stride=(1,1,1) & Spconv\_tensor\_1& 1600$\times$ 1408 $\times$ 41 $\times$ 16 \\
				\midrule
				Spconv\_tensor\_1  & 3DSparseCvonv, filter=(3,3,3), stride=(2,2,2) & Spconv\_tensor\_2& 800$\times$ 704 $\times$ 21 $\times$ 32 \\
				\midrule
				Spconv\_tensor\_2  & 3DSparseCvonv, filter=(3,3,3), stride=(2,2,2) & Spconv\_tensor\_3& 400$\times$ 352 $\times$ 11 $\times$ 64 \\
				\midrule
				Spconv\_tensor\_3  & 3DSparseCvonv, filter=(3,3,3), stride=(2,2,2) & Spconv\_tensor\_4& 200$\times$ 176 $\times$ 5 $\times$ 64 \\
				\hline
				Spconv\_tensor\_4  & 3DSparseCvonv, filter=(1,1,3), stride=(1,1,2) & Latent feature tensor& 200$\times$ 176 $\times$ 2 $\times$ 128\\
				\midrule
				\midrule
				\multicolumn{4}{c} {3D  Decoder} \\
				\midrule
				\midrule
				Latent feature tensor & 3DTransCvonv, filter=(3,3,3), stride=(2,2,2) & Dense\_tensor\_1& 400$\times$ 354 $\times$ 4 $\times$ 32 \\
				\midrule
				Dense\_tensor\_1  & 3DTransCvonv, filter=(3,3,3), stride=(2,2,4) & Dense\_tensor\_2& 800$\times$ 704 $\times$ 14 $\times$ 8 \\
				\midrule
				Dense\_tensor\_2  & 3DTransCvonv, filter=(3,3,3), stride=(2,2,3)& Occupied voxels $\textbf{P}$& 1600$\times$ 1408 $\times$ 41 $\times$ 1\\
				\bottomrule
			\end{tabular}
		}
		\label{tab:network}
	\end{table*}
	
	Given the instances from large-scale LiDAR point clouds, self-supervised pre-training is to train the network with the unlabeled data to generate representative features. Inspired by the excellent performance of masked autoencoding~\cite{bert,mae,point_mae}, we design the masked occupancy autoencoding network for 3D perception. The proposed Occupancy-MAE randomly masks the voxels and then reconstructs the occupancy values of voxels with an autoencoder network. The pretext task of occupancy prediction is trained with the binary cross-entropy loss. Occupancy-MAE covers most of the backbone networks, excluding the final head section.
	
	With the $n_s$ unlabeled point cloud data $\left \{\textbf{X}^i \right \}_{i=1}^{n_s}$, we aim to first pre-train the masked autoencoding network $\phi_{pre}$ to learn the high-level semantics. Then we use the pre-trained model to warm up the network $\phi_{s}$ of downstream tasks. We also extend the pre-training method to domain adaptative task on the target point clouds $\left \{\textbf{X}^j\right \}_{j=1}^{n_t}$. 
	Detailed information on the layers of Occupancy-MAE is listed in Table~\ref{tab:network}. The decoder consists of only two or three 3D Deconvolutional layers, making it quite lightweight.

	\subsection{Range-aware Random Masking}
	In this work, we adopt the common approach of dividing the LiDAR point clouds into spaced voxels, which is often used in 3D perception models~\cite{second,cylinder3d}. For LiDAR point clouds with dimensions $W\times H\times D$ along the $X\times Y\times Z$ axes, the size of each voxel is $v_W\times v_H\times v_D$, resulting in a total of $n_l$ voxels, of which $n_v$ contain points. Voxel-based methods are more computationally efficient than point-based methods~\cite{pointrcnn}, making them well-suited for processing large-scale LiDAR point clouds in applications such as self-driving cars.
	
	While the random masking strategy has been proven effective in pre-training models for language~\cite{bert}, images~\cite{mae}, and small-scale point clouds~\cite{point_mae}, the distribution of large-scale LiDAR point clouds is unique due to their sparsity levels being correlated with distance from the LiDAR sensor. Points close to the sensor are densely packed, while those further away are much more sparse. Therefore, we cannot apply the same masking strategy to both near-range and far-range points. Instead, we propose a range-aware random masking strategy that masks a small percentage of data for far-range points.
	
	To this end, we introduce the range-aware random masking strategy which takes into account the distance information. We adhere to the commonly employed distance configuration for evaluating LiDAR point clouds detection results~\cite{waymo,once}, dividing the occupied voxels into three groups based on their distance to the LiDAR sensor: 0-30 meters, 30-50 meters, and $>$50 meters. The masking ratio decreases as the distance increases, and we use a piecewise approach to apply the random masking strategy to each group. The corresponding numbers of voxels are $n_{v1}$, $n_{v2}$, and $n_{v3}$. We take the random masking strategy for each group with the descending masking ratio $r1$, $r2$, and $r3$ (i.e., $r1>r2>r3$). Thus the number of unmasked occupied voxels is $n_{un}=n_{v1}(1-r1)+n_{v2}(1-r2)+n_{v3}(1-r3)$, and the set of voxels $\textbf{V}_{input} \in R^{n_{un}\times 4}$ is used as training data. 
	The ground truth, denoted by $\textbf{T} \in \{0,1\}^{n_l \times 1}$, represents the occupancy status of the voxels. Each voxel can either contain points (occupied) or be empty (free). A value of 1 indicates an occupied voxel, while a value of 0 indicates a free voxel.
	
	It is noted that more division groups for range-aware masking are also applicable but will cause lots of preprocessing time in pre-training. Therefore, we
	struck a balance between accuracy and speed by implementing three groups of distance for range-aware masking. 
	\subsection{3D Sparse Convolutional Encoder}
	
	The Transformer network used in masked autoencoding for NLP~\cite{bert}, 2D vision~\cite{mae}, and small-scale point clouds~\cite{point_mae, maskpoint} performs self-attention on unmasked portions of the training data, which are unaffected by the masking. However, with millions of points in a 3D scene, even after masking 90\% of voxels, there are still hundreds of thousands of unmasked voxels, making it impractical for the Transformer network to aggregate information from such a vast amount of input data~\cite{bert}. Some algorithms (e.g., Voxel-MAE~\cite{voxelmae}, MV-JAR~\cite{mv-jar}) utilize Transformers for masking, but they are only applicable to 2D pillar structures, disregarding the 3D  height information. 
	
	To address this, the 3D Sparse Convolution~\cite{graham20183d, second} was proposed for processing large-scale point clouds. This approach uses positional encoding to aggregate information only from occupied voxels, resulting in high efficiency. Popular 3D perception methods~\cite{second, pv_rcnn, centerpoint, cylinder3d} have been developed using this technique. Motivated by this, we adopt the 3D Spatially Sparse Convolution from SECOND~\cite{second} to build our encoder network, allowing us to aggregate information only from unmasked, occupied voxels using the positional encoding module. Our voxel masking strategy thus reduces the memory complexity of training, similar to how the Transformer network works in NLP~\cite{bert}, 2D vision~\cite{mae}, and small-scale point clouds~\cite{point_mae, maskpoint, point_betr, pointm2ae}.
	
	\subsection{Lightweight 3D Decoder}
	
	Our decoder is composed of 3D Deconvolutional layers, with the last layer outputting the probability of each voxel containing points, resulting in an output tensor $\textbf{P}\in R^{n_l\times 1}$. During pre-training, the decoder's only purpose is to perform occupied voxel reconstruction. By shifting the masked tokens to the decoder, we encourage the encoder to learn better latent features for downstream tasks. The decoder is lightweight, consisting of only two or three 3D Deconvolutional layers, making it scalable to larger perception ranges.

	\subsection{Reconstructed Occupancy Target}
	
	The main objective of most masked autoencoding works is to reconstruct the masked parts through a regression task~\cite{point_mae,point_betr,pointm2ae}, which is not challenging for the network due to the positional encoding of voxels. However, in 3D perception, the occupancy structure of the general 3D world plays a crucial role in perception models~\cite{voxelnet,second,cylinder3d}. For instance, Tesla introduced the Occupancy Networks for autonomous driving~\cite{tesla}. Motivated by this, we propose the occupancy prediction task for large-scale outdoor LiDAR point clouds pre-training, aiming to encourage the network to reason over high-level semantics to recover the masked occupancy distribution of the 3D scene from a small number of visible voxels. 
	Considering the significant presence of empty voxels, the prediction of occupancy presents a binary classification challenge with class imbalance. To address this, we employ focal loss for binary occupancy classification, using the predicted occupancy values $\textbf{P}$ and the ground truth occupied voxels $\textbf{T}$.
	
	
	\begin{equation} \label{loss}
	loss = -\frac{1}{\text{batch}}\frac{1}{n_l }\sum_{i=1}^{\text{batch}}\sum_{j=1}^{n_l}\alpha_t \left(1 - \textbf{P}_t^{ij}\right)^\gamma \log(\textbf{P}_t^{ij}),
	\end{equation}
	where $\textbf{P}^{ij}$ represents the predicted probability of voxel $j$ in the $i$-th training sample and batch corresponds to the batch size. The weighting factor $\alpha$ for positive/negative examples is set to 2, and the weighting factor $\gamma$ for easy/hard examples is 0.25. For class 1, $\alpha_t=\alpha$ and $\textbf{P}_t^{ij} = \textbf{P}^{ij}$. For class 0, $\alpha_t=1-\alpha$ and $\textbf{P}_t^{ij} = 1-\textbf{P}^{ij}$.
	
	\subsection{Compared with Existing Masked LiDAR Methods}
	Recently, several self-supervised pre-training algorithms based on masked autoencoders have been proposed for large-scale outdoor LiDAR point clouds, including Voxel-MAE~\cite{voxelmae}, MV-JAR~\cite{mv-jar}, GeoMAE~\cite{geomae}, and GD-MAE~\cite{gd-mae}) have been proposed.
	
	Our method differs from these existing approaches in four key aspects. Firstly, the above masked LiDAR methods are limited to pillar-based methods, which overlook the crucial height information of the point clouds, while our method is applicable to both voxel-based and pillar-based 3D object detection (e.g., VoxelNet~\cite{voxelnet} and PointPillars~\cite{pointpillars}), 3D segmentation (e.g., Cylinder3D~\cite{cylinder3d}), and domain adaptation algorithms (e.g., ST3D~\cite{st3d}). Secondly, we introduce the range-aware masking strategy, which treats point clouds at different distances differently, accounting for varying information densities.
	Thirdly, instead of regressing the LiDAR point clouds directly, which may not be appropriate due to the input points having positions (x, y, z) and offering a shortcut for regression, we design a 3D occupancy prediction objective for the entire surrounding scene. In contrast, GeoMAE~\cite{geomae} and MV-JAR~\cite{mv-jar} use a 2D occupancy prediction target, neglecting the crucial information along the z-axis.
	Fourthly, our method utilizes sparse 3D Convolutions, enabling it to focus on 3D voxels and effectively incorporate volumetric information. In comparison, the above masked LiDAR methods use Transformers, which only aggregate 2D pillar features and disregard height information.
	
	\section{Experiments}
	\begin{table*}[h]
		\centering
		\caption{Quantitative detection performance achieved by different 
			self-supervised learning methods on the ONCE \emph{val} set. The pre-training process is on the unlabeled small set that contains 100k scenes.}
		\resizebox{\textwidth}{!}{
			{
				\begin{tabular}{c|cccc|cccc|cccc|c}
					\toprule
					\multirow{2}*{\textbf{Method}} &\multicolumn{4}{c|}{\textbf{Vehicle}} &\multicolumn{4}{c|}{\textbf{Pedestrian}}  &\multicolumn{4}{c|}{\textbf{Cyclist}}&\multirow{2}*{\textbf{mAP}$\uparrow$}\\
					&\textbf{overall}$\uparrow$&\textbf{0-30m}&\textbf{30-50m}&\textbf{50m-inf}&\textbf{overall}$\uparrow$&\textbf{0-30m}&\textbf{30-50m}&\textbf{50m-inf}&\textbf{overall}$\uparrow$&\textbf{0-30m}&\textbf{30-50m}&\textbf{50m-inf}& \\
					\midrule
					SECOND~\cite{second} &71.19&84.04&63.02&47.25&26.44&29.33&24.05&18.05&58.04&69.96&52.43&34.61&51.89 \\
					\midrule
					BYOL~\cite{byol}  &68.02&	81.01&	60.21&	44.17&	19.50&	22.16&	16.68&	12.06&	50.61	&62.46&	44.29&	28.18&	$46.04^{\textcolor{teal} {-5.85}}$ \\
					ProposalContrast~\cite{proposalcontrast} &72.99&	\textbf{84.41}&	65.92&	50.11&	25.77&	27.95&	23.74&	18.06&	58.23&	69.99&	53.03&	35.48&	$52.33^{\textcolor{teal} {+0.44}}$\\
					\midrule
					SwAV~\cite{swav}&72.71&	83.68&	65.91&	50.10&	25.13&	27.77&	22.77&	16.36&	58.05&	69.99&	52.23&	34.86&	$51.96^{\textcolor{teal} {+0.07}}$ \\
					DeepCluster~\cite{deepcluster} &\textbf{73.19}&84.25&	\textbf{66.86}&	\textbf{50.47}&	24.00&	26.36&	21.73&	\textbf{16.79}&	\textbf{58.99}&	\textbf{70.80}&	\textbf{53.66}&	\textbf{36.17}&	$52.06^{\textcolor{teal} {+0.17}}$ \\
					\midrule
					Occupancy-MAE &72.78 &83.77&66.01&50.26& \textbf{27.49} &\textbf{30.54}&\textbf{25.28}&16.11&57.26&69.71&52.31&33.51&${\bf 52.51}^{\textcolor{teal} {+0.62\%}}$ \\
					\bottomrule
				\end{tabular}
		}}
		\label{tab:ONCE_small}
	\end{table*}
	\subsection{Dataset}
	
	The four popular autonomous driving datasets: ONCE~\cite{once}, KITTI~\cite{kitti}, Waymo~\cite{waymo}, and nuScenes~\cite{nuscenes} are used in our experiments. 
	
	$\textbf{ONCE}$. The ONCE dataset~\cite{once } comprises 1 million LiDAR scenes and 7 million corresponding camera images. The data encompasses diverse areas, time periods, and weather conditions, providing a comprehensive representation of real-world driving scenarios. Notably, one of the key objectives of the ONCE dataset is to facilitate research that explores the potential of utilizing large-scale unlabeled data, opening up opportunities for innovative approaches in autonomous driving research and development.
	
	$\textbf{KITTI}$. The KITTI dataset\cite{kitti} is one of the most popular autonomous driving datasets, which provides 7,481 training samples and 7,518 testing samples. The 3D bounding box annotations are only provided within the Field of View (FoV) of the front camera. We follow the common 50/50 \emph{train}/\emph{val} split and use the official KITTI evaluation metrics for three-level evaluation (Easy, Moderate, Hard) and the mean average precision is evaluated. 
	
	$\textbf{Waymo Open Dataset}$. The Waymo Open Dataset~\cite{waymo} is a recently released large-scale autonomous driving dataset, which consists of total of 798 training sequences with around 158,361 LiDAR samples, and 202 validation sequences with 40,077 LiDAR samples. Following the popular point cloud detection codebase OpenPCDet~\cite{openpcdet}, we subsample a single frame of 20\% data (about 32k frames) of all the training samples as the training set. It annotated the objects in the full 360$^{\circ}$ field. The official evaluation metrics of Waymo are mean average precision (AP) and mean average precision weighted by heading (APH) in difficulty levels (L1 and L2).
	
	$\textbf{nuScenes}$. The nuScenes dataset~\cite{nuscenes} is another popular autonomous driving dataset. There are a total of 28,130 training samples and 6,019 validation samples. We evaluate with the official evaluation metrics of the nuScenes Detection Score (NDS), mean average precision (mAP), average translation error (ATE), average scale error (ASE), average orientation error (AOE), average velocity error (AVE), average attribute error (AAE). 
	
	\begin{table*}[t]
		\begin{minipage}[c]{0.5\textwidth}
			\caption{Performance comparison on the KITTI test split evaluated by the mean Average Precision with 40 recall positions at a moderate difficulty level.}
			\centering
			\resizebox{0.99\textwidth}{!}{
				\begin{tabular}{c|c|c|c|c}
					\toprule
					\multirow{2}*{\textbf{Method}}&\multicolumn{2}{c|}{\textbf{Car}} &\multicolumn{2}{c}{\textbf{Cyclist}} \\
					\cmidrule{2-5}
					&\textbf{3D}$\uparrow$&\textbf{BEV}$\uparrow$&\textbf{3D}$\uparrow$&\textbf{BEV}$\uparrow$\\
					\midrule
					\footnotesize{SECOND~\cite{second}} &72.55&83.77&52.08 &56.05  \\
					\footnotesize{Occupancy-MAE + SECOND} &${\bf 72.87}^{\textcolor{teal} {+0.32 }}$&${\bf 83.96}^{\textcolor{teal} {+0.19}}$&${\bf 54.84}^{\textcolor{teal} {+1.76}}$&${\bf60.67}^{\textcolor{teal} {+4.62}}$ \\
					\bottomrule
			\end{tabular}}
			\label{tab:kitti_test}
		\end{minipage}
		\begin{minipage}[c]{0.5\textwidth}
			\caption{Performance comparison on the KITTI \emph{val} split evaluated by the mean Average Precision with 40 recall positions.}
			\centering
			\resizebox{0.8\textwidth}{!}{
				\begin{tabular}{c|c|c|c}
					\toprule
					\textbf{Method} &\textbf{Car}$\uparrow$ &\textbf{Pedestrian}$\uparrow$ &\textbf{Cyclist}$\uparrow$\\
					\midrule
					\footnotesize{SST~\cite{sst}} &81.46&46.52&65.59 \\
					\footnotesize{Voxel-MAE~\cite{voxelmae} + SST} &$81.53^{\textcolor{teal} {+0.07}}$&$46.63^{\textcolor{teal} {+0.11}}$&$65.72^{\textcolor{teal} {+0.13}}$\\
					\footnotesize{MV-JAR~\cite{mv-jar} + SST} &$81.75^{\textcolor{teal} {+0.29}}$&$47.56^{\textcolor{teal} {+1.04}}$&$66.37^{\textcolor{teal} {+0.78}}$   \\
					\footnotesize{GeoMAE~\cite{geomae} + SST} &$81.84^{\textcolor{teal} {+0.38}}$&$48.18^{\textcolor{teal} {+1.66}}$&$67.01^{\textcolor{teal} {+1.42}}$   \\
					\footnotesize{GD-MAE~\cite{gd-mae} + SST} &$82.01^{\textcolor{teal} {+0.55}}$&$48.40^{\textcolor{teal} {+1.88}}$&$67.16^{\textcolor{teal} {+1.57}}$ \\
					\footnotesize{Occupancy-MAE + SST} &${\bf 82.24}^{\textcolor{teal} {+0.78}}$&${\bf 48.95}^{\textcolor{teal} {+2.43 }}$&${\bf 67.78}^{\textcolor{teal} {+2.19 }}$ \\
					\bottomrule
			\end{tabular}}
			\label{tab:kitti_pillars}
		\end{minipage}
		
	\end{table*}
	
	%

	\begin{table*}[ht]
		\centering
		\caption{Performance comparison on the KITTI \emph{val} split evaluated by the AP with 40 recall positions at moderate difficulty level.}
		\resizebox{\textwidth}{!}
		{
			{
				\begin{tabular}{c|c|c|c|c|c|c|c}
					\toprule
					\textbf{Method} &\textbf{Car} $\uparrow$&\textbf{Pedestrian}$\uparrow$ &\textbf{Cyclist}$\uparrow$&\textbf{Method} &\textbf{Car}$\uparrow$ &\textbf{Pedestrian}$\uparrow$ &\textbf{Cyclist}$\uparrow$ \\
					\midrule
					SECOND [45] &81.50 &48.82 &65.72 &PV-RCNN [31] &84.50 &57.06 &70.14 \\
					ALSO~\cite{also} + SECOND &$81.97^{\textcolor{teal} {+0.47}}$ &$51.93^{\textcolor{teal} {+3.11}}$ &${\bf69.14}^{\textcolor{teal} {+3.42}}$ &ALSO~\cite{also} + PV-RCNN &$84.72^{\textcolor{teal} {+0.22}}$ &$58.49^{\textcolor{teal} {+1.43}}$ &$75.06^{\textcolor{teal} {+4.92}}$ \\
					Occupancy-MAE + SECOND &${\bf 81.98}^{\textcolor{teal} {+0.48}}$&${\bf 53.67}^{\textcolor{teal} {+4.85}}$& $69.08^{\textcolor{teal} {+3.36}}$&Occupancy-MAE + PV-RCNN &${\bf84.82}^{\textcolor{teal} {+0.32}}$&${\bf59.07}^{\textcolor{teal} {+2.01}}$&${\bf75.68}^{\textcolor{teal} {+5.54}}$ \\
					
					\bottomrule
				\end{tabular}
		}}
		\label{tab:kitti}
	\end{table*}
	
	\begin{table*}[ht]
		\centering
		\caption{Performance comparison on the KITTI \emph{val} split with AP calculated by 11 recall positions evaluating bounding box and orientation.}
		\renewcommand\arraystretch{1.2}  
		\resizebox{\textwidth}{!}{
			{
				\begin{tabular}{c|c|ccc|ccc|ccc}
					\toprule
					\multirow{2}*{\textbf{Evaluation}} &\multirow{2}*{\textbf{Method}}&\multicolumn{3}{c|}{\textbf{Car}}&\multicolumn{3}{c|}{\textbf{Pedestrian}}&\multicolumn{3}{c}{\textbf{Cyclist}} \\
					&&\textbf{Easy}$\uparrow$ &\textbf{Moderate}$\uparrow$ &\textbf{Hard}$\uparrow$&\textbf{Easy}$\uparrow$ &\textbf{Moderate}$\uparrow$ &\textbf{Hard}$\uparrow$&\textbf{Easy}$\uparrow$ &\textbf{Moderate}$\uparrow$ &\textbf{Hard}$\uparrow$ \\
					\midrule
					\multirow{2}*{bbox}&SECOND [45] &90.73&89.76&88.94 &68.70& 65.27&62.52&87.88&75.43&71.67\\
					\cline{2-11}
					&Occupancy-MAE + SECOND &${\bf94.81}^{\textcolor{teal} {+4.08}}$&${\bf89.98}^{\textcolor{teal} {+0.22}}$&${\bf89.35}^{\textcolor{teal} {+0.41}}$&${\bf70.37}^{\textcolor{teal} {+1.67}}$& ${\bf67.45}^{\textcolor{teal} {+2.18}}$&${\bf65.14}^{\textcolor{teal} {+2.62}}$&${\bf91.82}^{\textcolor{teal} {+3.94}}$&${\bf78.65}^{\textcolor{teal} {+3.22}}$&${\bf73.77}^{\textcolor{teal} {+2.10}}$\\
					\midrule
					\multirow{2}*{aos}&SECOND [45]&90.73&89.63&88.70&63.46&60.13&56.93&87.63&74.67 &71.00  \\
					\cline{2-11}
					&Occupancy-MAE + SECOND &${\bf94.66}^{\textcolor{teal} {+3.93}}$&${\bf89.88}^{\textcolor{teal} {+0.25}}$&${\bf88.92}^{\textcolor{teal} {+0.22}}$&${\bf65.33}^{\textcolor{teal} {+1.97}}$&${\bf61.55}^{\textcolor{teal} {+1.42}}$&${\bf59.23}^{\textcolor{teal} {+2.30}}$&${\bf91.57}^{\textcolor{teal} {+1.94}}$&${\bf78.42}^{\textcolor{teal} {+3.75}}$& ${\bf73.50}^{\textcolor{teal} {+2.50}}$\\
					\bottomrule
				\end{tabular}
		}}
		\label{tab:kitti_bbox}
	\end{table*}

	\begin{table*}[t]
		\centering
		\caption{Quantitative detection performance on the Waymo \emph{val} set. The models are trained on 20\% Waymo training set.}
		\resizebox{\textwidth}{!}
		{
			{
				\begin{tabular}{c|cc|cc|cc|cc|cc|cc}
					\toprule
					\multirow{2}*{\textbf{Method}} &\multicolumn{2}{c|}{\textbf{Vec\_L1}} &\multicolumn{2}{c|}{\textbf{Vec\_L2}}  &\multicolumn{2}{c|}{\textbf{Ped\_L1}}&\multicolumn{2}{c|}{\textbf{Ped\_L2}}&\multicolumn{2}{c|}{\textbf{Cyc\_L1}} &\multicolumn{2}{c}{\textbf{Cyc\_L2}}\\
					&\textbf{AP}$\uparrow$&\textbf{APH}$\uparrow$&\textbf{AP}$\uparrow$&\textbf{APH}$\uparrow$&\textbf{AP}$\uparrow$&\textbf{APH}$\uparrow$&\textbf{AP}$\uparrow$&\textbf{APH}$\uparrow$&\textbf{AP}$\uparrow$&\textbf{APH}$\uparrow$&\textbf{AP}$\uparrow$&\textbf{APH}$\uparrow$ \\
					\midrule
					CenterPoint~\cite{centerpoint} &71.33&70.76&63.16&62.65&72.09&65.49	&64.27&58.23&68.68&67.39&66.11&64.87 \\
					GCC-3D~\cite{gcc3d} +  CenterPoint&-&-&$63.97^{\textcolor{teal} {+0.81}}$&$63.47^{\textcolor{teal} {+0.82}}$&-&-&$64.23^{\textcolor{teal} {-0.04}}$&$58.47^{\textcolor{teal} {+0.24}}$&-&-&$67.68^{\textcolor{teal} {+1.57}}$&$66.44^{\textcolor{teal} {+1.57}}$ \\
					Occupancy-MAE + CenterPoint&${\bf71.89}^{\textcolor{teal} {+0.56}}$&${\bf71.33}^{\textcolor{teal} {+0.57}}$&${\bf64.05}^{\textcolor{teal} {+0.89}}$&${\bf63.53}^{\textcolor{teal} {+0.88}}$&${\bf73.85}^{\textcolor{teal} {+1.76}}$&${\bf67.12}^{\textcolor{teal} {+1.63}}$&${\bf65.78}^{\textcolor{teal} {+1.51}}$&${\bf59.62}^{\textcolor{teal} {+1.39}}$&${\bf70.29}^{\textcolor{teal} {+1.61}}$&${\bf69.03}^{\textcolor{teal} {+1.64}}$&${\bf67.76}^{\textcolor{teal} {+1.65}}$&${\bf66.53}^{\textcolor{teal} {+1.66}}$ \\
					\midrule
					PV-RCNN~\cite{pv_rcnn} &75.41&74.74&67.44&66.80&71.98&61.24	&63.70&53.95&65.88&64.25&63.39&61.82 \\
					Occupancy-MAE + PV-RCNN& ${\bf75.94}^{\textcolor{teal} {+9.53}}$&${\bf75.28}^{\textcolor{teal} {+0.54}}$&${\bf67.94}^{\textcolor{teal} {+0.50}}$&${\bf67.34}^{\textcolor{teal} {+0.54}}$&${\bf74.02}^{\textcolor{teal} {+2.04}}$&${\bf63.48}^{\textcolor{teal} {+2.24}}$&${\bf64.91}^{\textcolor{teal} {+1.21}}$&${\bf55.57}^{\textcolor{teal} {+1.62}}$&${\bf67.21}^{\textcolor{teal} {+1.33}}$&${\bf65.49}^{\textcolor{teal} {+1.24}}$&${\bf64.62}^{\textcolor{teal} {+1.23}}$&${\bf63.02}^{\textcolor{teal} {+1.20}}$ \\
					\midrule
				\end{tabular}
		}}
		\label{tab:waymo}
	\end{table*}
	
	\begin{table*}[t]
		\centering
		\caption{Quantitative detection performance on the Waymo \emph{val} set. The models are trained on 100\% Waymo training set.}
		\resizebox{\textwidth}{!}
		{
			{
				\begin{tabular}{c|cc|cc|cc|cc|cc|cc}
					\toprule
					\multirow{2}*{\textbf{Method}} &\multicolumn{2}{c|}{\textbf{Vec\_L1}} &\multicolumn{2}{c|}{\textbf{Vec\_L2}}  &\multicolumn{2}{c|}{\textbf{Ped\_L1}}&\multicolumn{2}{c|}{\textbf{Ped\_L2}}&\multicolumn{2}{c|}{\textbf{Cyc\_L1}} &\multicolumn{2}{c}{\textbf{Cyc\_L2}}\\
					&\textbf{AP}$\uparrow$&\textbf{APH}$\uparrow$&\textbf{AP}$\uparrow$&\textbf{APH}$\uparrow$&\textbf{AP}$\uparrow$&\textbf{APH}$\uparrow$&\textbf{AP}$\uparrow$&\textbf{APH}$\uparrow$&\textbf{AP}$\uparrow$&\textbf{APH}$\uparrow$&\textbf{AP}$\uparrow$&\textbf{APH}$\uparrow$ \\
					\midrule
					SECOND~\cite{second} &72.27&71.69&63.85&63.33&68.70&58.18&60.72&51.31	&60.62&59.28&58.34&57.05 \\
					ProposalContrast~\cite{proposalcontrast} + SECOND &-&-&$64.50^{\textcolor{teal} {+0.65}}$&$63.90^{\textcolor{teal} {+0.57}}$&-&-&$60.33^{\textcolor{teal} {-0.39}}$&$51.00^{\textcolor{teal} {-0.31}}$	&-&-&$57.90^{\textcolor{teal} {-0.44}}$&$56.60^{\textcolor{teal} {-0.45}}$ \\
					BEV-MAE~\cite{bevmae} + SECOND &-&-&$64.42^{\textcolor{teal} {+0.57}}$&$63.87^{\textcolor{teal} {+0.54}}$&-&-&$59.97^{\textcolor{teal} {-0.75}}$&$50.65^{\textcolor{teal} {-0.66}}$	&-&-&$58.69^{\textcolor{teal} {+0.35}}$&$57.39^{\textcolor{teal} {+0.34}}$ \\
					Occupancy-MAE + SECOND &${\bf73.68}^{\textcolor{teal} {+1.41}}$&${\bf73.12}^{\textcolor{teal} {+1.43}}$&${\bf65.72}^{\textcolor{teal} {+1.97}}$&${\bf65.19}^{\textcolor{teal} {+1.86}}$&${\bf70.21}^{\textcolor{teal} {+1.51}}$&${\bf60.45}^{\textcolor{teal} {+2.27}}$&${\bf62.52}^{\textcolor{teal} {+1.80}}$&${\bf53.69}^{\textcolor{teal} {+2.38}}$&${\bf63.13}^{\textcolor{teal} {+2.51}}$&${\bf61.88}^{\textcolor{teal} {+2.60}}$&${\bf60.83}^{\textcolor{teal} {+2.49}}$&${\bf59.63}^{\textcolor{teal} {+2.58}}$ \\
					\bottomrule
				\end{tabular}
		}}
		\label{tab:waymo100}
	\end{table*}
	
	\begin{table*}[h]
		\centering
		\caption{Quantitative detection performance achieved by different methods on the nuScenes \emph{val} set.}
		{
			{
				\begin{tabular}{c|c|c|c|c|c|c|c|c}
					\toprule
					\textbf{Method}&\textbf{Frames} &\textbf{mAP}$\uparrow$&\textbf{NDS}$\uparrow$&\textbf{mATE}$\downarrow$&\textbf{mASE}$\downarrow$&\textbf{mAOE}$\downarrow$&\textbf{mAVE}$\downarrow$&\textbf{mAAE}$\downarrow$ \\
					\midrule
					CenterPoint~\cite{centerpoint} &-&56.0&64.5&30.1&25.6& 38.3&21.9	&18.9	 \\
					GCC-3D~\cite{gcc3d} + CenterPoint &-&$57.3^{\textcolor{teal} {+1.3}}$&$65.0^{\textcolor{teal} {+0.5}}$&-&-&-&-	&-	 \\
					BEV-MAE~\cite{bevmae} + CenterPoint &-&$57.2^{\textcolor{teal} {+1.2}}$&$65.1^{\textcolor{teal} {+0.6}}$&-&-&-&-	&-	 \\
					Occupancy-MAE + CenterPoint &1&$ 56.5^{\textcolor{teal} {+0.5}}$&$ 65.0^{\textcolor{teal} {+0.5}}$& 29.7 & 25.2&38.4& 21.5&18.7\\
					Occupancy-MAE + CenterPoint &3&${\bf 57.8}^{\textcolor{teal} {+1.8}}$&${\bf 65.9}^{\textcolor{teal} {+1.4}}$&{\bf 28.8} &{\bf 24.4}&{\bf37.5}&{\bf 20.6}&{\bf 18.0}\\
					\bottomrule
				\end{tabular}
		}}
		\label{tab:nuscenes}
	\end{table*}
	
	\begin{table*}[t]
		\centering
		\caption{Quantitative segmentation performance achieved by different methods on the nuScenes \emph{val} set.}
		\resizebox{\textwidth}{!}{
			{
				\begin{tabular}{c|c|c|c|c|c|c|c|c|c|c|c|c|c|c|c|c|c|c}
					\toprule
					\textbf{Method} &\textbf{Epoch}&\textbf{mIoU}$\uparrow$ &\textbf{\rotatebox{90}{barrier}}&\textbf{\rotatebox{90}{bicycle}}&\textbf{\rotatebox{90}{bus}}&\textbf{\rotatebox{90}{car}}&\textbf{\rotatebox{90}{construction}}&\textbf{\rotatebox{90}{motorcycle}}&\textbf{\rotatebox{90}{pedestrian}}&\textbf{\rotatebox{90}{traffic-cone}}&\textbf{\rotatebox{90}{trailer}}&\textbf{\rotatebox{90}{truck}}&\textbf{\rotatebox{90}{driveable}}&\textbf{\rotatebox{90}{others}}&\textbf{\rotatebox{90}{sidewalk}}&\textbf{\rotatebox{90}{terrain}}&\textbf{\rotatebox{90}{manmade}}&\textbf{\rotatebox{90}{vegetation}}\\
					\midrule
					\multirow{2}*{Cylinder3D~\cite{cylinder3d}} &15 &70.22&71.5&16.7&88.2&85.2&40.5&71.5&68.5&61.1&56.7&79.9&96.0&{\bf70.9}&72.3&73.2&86.2&85.3\\
					&25 &70.83&72.5&17.1&88.2&{\bf85.3}&{\bf46.0}&71.3&73.7&58.7&{\bf61.9}&{\bf81.0}&96.0&66.0&71.9&71.6&86.6&85.4\\
					\midrule
					\multirow{2}*{Occupancy-MAE + Cylinder3D}&15 &$71.61^{\textcolor{teal} {+1.39}}$&72.6&29.3&{\bf89.4}&84.3&40.4&77.7&73.4&60.1&58.2&77.0&96.0&69.9&72.2&{\bf73.7}&86.3&{\bf85.5}\\
					&25 &${\bf72.85}^{\textcolor{teal} {+2.02}}$&{\bf74.6}&{\bf33.0}&89.3&85.2&40.9&{\bf78.6}&{\bf75.9}&{\bf62.9}&61.5&79.3&{\bf96.1}&70.5&{\bf72.7}&72.7&{\bf87.2}&85.4\\
					\bottomrule
				\end{tabular}
		}}
		\label{tab:nuscenes_seg}
	\end{table*}

	\begin{table*}[h]
		\centering
		\caption{Quantitative multi-object tracking performance achieved by different methods on the nuScenes \emph{val} set.}
		{
			{
				\begin{tabular}{c|c|c|c|c|c|c|c|c|c|c}
					\toprule
					\textbf{Tracker}&\textbf{Detector} &\textbf{AMOTA}$\uparrow$&\textbf{AMOTP}$\downarrow$&\textbf{MT}$\uparrow$&\textbf{ML}$\downarrow$&\textbf{TP}$\uparrow$&\textbf{FP}$\downarrow$&\textbf{FN}$\downarrow$&\textbf{IDS}$\downarrow$&\textbf{FRAG}$\downarrow$ \\
					\midrule
					\multirow{4}*{GNN-PMB~\cite{gnn-pmb}} &PointPillars~\cite{pointpillars}&0.311 &1.231 &2754 &2236 &60929 &${\bf9993}$ &39945 &1023 &769	 \\
					&Occupancy-MAE + PointPillars & ${\bf0.324}^{\textcolor{teal} {+1.3}}$ &${\bf1.121}^{\textcolor{teal} {+1.0}}$ &${\bf2912}$ &${\bf2089}$ &${\bf65241}$ &10117 &${\bf38083}$ &${\bf906}$ &${\bf698}$	 \\
					&CenterPoint~\cite{centerpoint} &0.707 &0.560 &4608 &1347 &83134 &12362 &18113 &650 &345	 \\
					& Occupancy-MAE + CenterPoint &${\bf0.719}^{\textcolor{teal} {+1.2}}$ &${\bf0.548}^{\textcolor{teal} {+1.2}}$ &${\bf4801}$ &${\bf1211}$ &${\bf84251}$ &${\bf10152}$ &${\bf16993}$ &${\bf498}$ &${\bf284}$\\
					\bottomrule
				\end{tabular}
		}}
		\label{tab:mot}
	\end{table*}
	
	\begin{table}[t]
		\centering
		\caption{Quantitative results of adaptation tasks. Our Occupancy-MAE improves the performance of UDA 3D object detection.}
		\resizebox{0.48\textwidth}{!}{
			\begin{tabular}{c|c|cc}
				\toprule
				\multirow{2}*{\textbf{Task}} &\multirow{2}*{\textbf{Method}}&\multicolumn{2}{c}{\textbf{PV-RCNN}} \\  
				&&\textbf{BEV}$\uparrow$ &\textbf{3D}$\uparrow$\\
				\midrule
				\multirow{6}*{Waymo $\rightarrow$ KITTI}
				& Oracle&88.98 &82.50 \\ 
				& Source-only &61.18 &22.01 \\ 
				& SN~\cite{sn}  &79.78 & 63.60\\ 
				& ST3D~\cite{st3d}  &84.10 & 64.78\\  
				& ST3D(w/SN)~\cite{st3d}&86.65 & 76.86\\ 
				& Occupancy-MAE + ST3D&${\bf 85.52}^{\textcolor{teal} {+1.42}}$ & ${\bf 65.24}^{\textcolor{teal} {+0.46}}$ \\ 
				& Occupancy-MAE + ST3D(w/SN)&${\bf 87.15}^{\textcolor{teal} {+0.50}}$ &${\bf  77.13}^{\textcolor{teal} {+0.27}}$ \\ 
				\midrule
				\multirow{6}*{nuScenes $\rightarrow$ KITTI}
				& Oracle&88.98 & 82.50 \\ 
				& Source-only &68.15 & 37.17 \\ 
				& SN~\cite{sn} &60.48 & 49.47\\ 
				& ST3D~\cite{st3d}&78.36 & 70.85\\ 
				& ST3D(w/SN)~\cite{st3d} &84.29 & 72.94\\ 
				& Occupancy-MAE + ST3D &${\bf 78.66}^{\textcolor{teal} {+0.30}}$ &${\bf  71.24}^{\textcolor{teal} {+0.39}}$ \\ 
				& Occupancy-MAE + ST3D(w/SN)&${\bf 85.43}^{\textcolor{teal} {+1.14}}$ & ${\bf 73.22}^{\textcolor{teal} {+0.28}}$ \\ 
				\bottomrule
			\end{tabular}
		}
		\label{tab:transfer}
	\end{table}
	
	\subsection{Experimental Setup}
	We evaluate the effectiveness of our proposed model by conducting three downstream tasks on four autonomous driving datasets~\cite{once,kitti,waymo,nuscenes}. To implement 3D object detection and unsupervised domain adaptation tasks, we utilize the popular point clouds detection codebase OpenPCDet~\cite{openpcdet} (version 0.5.2). For the 3D semantic segmentation task, we use the open-sourced Cylinder3D~\cite{cylinder3d} as the pre-trained backbone, which applies the cylindrical voxel partition. We first pre-train the Occupancy-MAE on the unlabeled raw set of the ONCE dataset and then fine-tune the perception models on the training set. However, for the KITTI, Waymo, and nuScenes datasets, pre-training and fine-tuning are both on the training set. The ONCE dataset provides a benchmark for self-supervised learning methods, but no codes are available. Thus, we only compare our Occupancy-MAE with self-supervised learning methods on the ONCE dataset.
	
	We utilize the pre-trained 3D encoder to initialize and warm up the backbones of the downstream tasks, without freezing the 3D encoder parameters during fine-tuning. Subsequently, we train the downstream tasks using the same training parameters as the original models. We set the masking ratio for voxels within 0-30 meters, 30-50 meters, and $>$ 50 meters to 90\%, 70\%, and 50\%, respectively. The number of pre-training epochs is 3. 
	For more detailed parameter setups, please refer to OpenPCDet~\cite{openpcdet, once} and Cylinder3D~\cite{cylinder3d}.
	
	\subsection{Results on Downstream Tasks}
	The experimental results have been validated multiple times with different seeds.
	\subsubsection{3D Object Detection}
	
	We begin by comparing our proposed method with several state-of-the-art self-supervised learning methods on the ONCE dataset~\cite{once} \emph{val} set. These include two contrastive learning methods ( BYOL~\cite{byol} and ProposalContrast~\cite{proposalcontrast}), two clustering-based methods (DeepCluster~\cite{deepcluster} and SwAV~\cite{swav}). Table~\ref{tab:ONCE_small} shows that our method, with the SECOND~\cite{second} detector as the pre-trained backbone, outperforms the self-supervised methods in terms of mAP. Our method achieves a 0.2\% $\sim$ 6\% performance gain over contrastive learning methods because the contrastive views of a 3D scene by data augmentations may be similar, causing the model to converge to a trivial solution~\cite{once}. Our Occupancy-MAE achieves comparable results with SwAV~\cite{swav} and DeepCluster~\cite{deepcluster}, with marginal improvement, but it is much simpler than clustering-based methods. 
	We attribute the performance gain of our method to the range-aware masking strategy and occupancy prediction task, which help to learn high-level semantic information for large-scale outdoor LiDAR point clouds.
	
	We present the performance of Occupancy-MAE on the KITTI test set in Table~\ref{tab:kitti_test}. Due to the limited submission times of 3 for KITTI policy, we only evaluate Occupancy-MAE with SECOND~\cite{second}. The KITTI test results of SECOND are reproduced by PV-RCNN~\cite{pv_rcnn} and the pedestrian class was not provided in PV-RCNN. Our method significantly improves 3D object detection results on the moderate level of cyclist class, from 52.08\% to 54.84\%, and on car class, from 72.55\% to 72.87\%. For bird-view detection of cyclist class, our approach outperforms training from scratch by increasing mAP by 4.62\% on moderate levels of cyclist class. 
	
	As shown in Table~\ref{tab:kitti}, Occupancy-MAE achieves better performance than training from scratch on the KITTI $\emph{val}$ set, especially for the small objects, i.e., pedestrian and cyclist classes, where the pre-trained model achieves about 1.5\% and 0.5\% gains, respectively. ALSO~\cite{also} shares similarities with our method as both involve predicting occupancy as an auxiliary task. However, our approach goes a step further by constructing a mask autoencoder framework, which enables the learning of more semantic information, thereby enhancing the precision of detection. We also observe that our method improves the evaluation of the bounding box and orientation in Table~\ref{tab:kitti_bbox}, indicating that Occupancy-MAE can effectively learn high-level semantic information from the masked voxels.
	
	Pillar-based methods, such as PointPillars~\cite{pointpillars} and SST~\cite{sst}, are variants of voxel-based methods with high computational efficiency. Pillar-based methods process point clouds in vertical columns (pillars), which may contain points of the road. We only consider the points above the road to build the occupancy labels. In Table~\ref{tab:kitti_pillars}, we show that Occupancy-MAE improves training from scratch by approximately 1\% $\sim$ 2\% performance gains on the pedestrian and car classes. Additionally, Occupancy-MAE outperforms Voxel-MAE~\cite{voxelmae}, MV-JAR~\cite{mv-jar}, GeoMAE~\cite{geomae}, and GD-MAE~\cite{gd-mae}, as the target of Occupancy-MAE (occupancy structure above the road) is more reasonable than regression of the points and pillar occupancy of Voxel-MAE~\cite{voxelmae}, MV-JAR~\cite{mv-jar}, GeoMAE~\cite{geomae}, and GD-MAE~\cite{gd-mae}. Besides, They are limited to pillar-based methods, whereas our Occupancy-MAE offers a wider range of applicability, treats points differently and reconstructs the overall 3D occupancy distribution.
	
	We evaluate Occupancy-MAE on the Waymo $\emph{val}$ set and present the results with 20\% training data in Table~\ref{tab:waymo} and with 100\% training data in Table~\ref{tab:waymo100}. Our pre-trained model outperforms training from scratch, especially for small objects, achieving approximately 1\% $\sim$ 2\% performance gains on the pedestrian and cyclist classes. Small objects contain only a few points, making them challenging to detect. The masked occupancy reconstruction task enables the network to fill in missing occupancy information, thereby improving the detection of small objects. We can see that pre-training with Occupancy-MAE leads to consistent performance gains over training from scratch, with around 1\% mAP improvement with 20\% training data and 2\% mAP improvement with 100\% training data. This further confirms that our method benefits from more self-supervised pre-training data.
	
	Lastly, we verify the effectiveness of Occupancy-MAE on nuScenes dataset and report the results in Table~\ref{tab:nuscenes}. The performance improvement of our Occupancy-MAE on nuScenes (32-beam) is limited because the point clouds are sparser than those in KITTI and Waymo (64-beam). Hence, we perform fusion on multiple frames of point clouds to construct dense occupancy labels. The results demonstrate that the fusion of 3 frames of point clouds yields an improvement of 1.8 in mAP and 1.4 in NDS. This suggests that the utilization of multi-frame fusion enhances the model's ability to learn contextual information about the surrounding scene, as it provides valuable insights into the overall structural characteristics of objects beyond the surface-level information offered by single-frame point clouds. We report results for SECOND~\cite{second}, CenterPoint~\cite{centerpoint}, and PV-RCNN~\cite{pv_rcnn} from OpenPCDet~\cite{openpcdet}, BEV-MAE~\cite{bevmae}, and ONCE~\cite{once}.
	
	\subsubsection{3D Semantic Segmentation}
	
	\begin{table}[t]
		\caption{Comparison of different pretext tasks.}
		\resizebox{0.48\textwidth}{!}{
			{
				\begin{tabular}{c|c|c|c}
					\toprule
					\textbf{Target} &\textbf{Car}$\uparrow$ &\textbf{Pedestrian}$\uparrow$ &\textbf{Cyclist}$\uparrow$ \\
					\midrule
					PV-RCNN~\cite{pv_rcnn} &83.61&57.90&70.47 \\
					Features Regression &$83.60^{\textcolor{teal} {-0.01}}$&$57.92^{\textcolor{teal} {+0.02}}$&$70.52^{\textcolor{teal} {+0.05}}$ \\
					Occupancy Classification &${\bf83.82}^{\textcolor{teal} {+0.21}}$&${\bf59.37}^{\textcolor{teal} {+1.47}}$&${\bf71.99}^{\textcolor{teal} {+1.52}}$ \\
					\bottomrule
				\end{tabular}
		}}
		\label{tab:task}
	\end{table}
	
	\begin{table}[t]
		\centering
		\caption{Impacts of pre-training times.}
		\setlength{\tabcolsep}{5.3mm}
		{
			\begin{tabular}{c|c|c|c}
				\toprule
				\textbf{Epoch} &\textbf{Easy}$\uparrow$ &\textbf{Moderate}$\uparrow$ &\textbf{Hard}$\uparrow$ \\
				\midrule
				2 &81.26&67.23&61.68  \\
				\midrule
				3 &82.09&{\bf 68.05}&{\bf 62.08} \\
				\midrule
				4 &{\bf 82.12}&68.00&61.99 \\
				\bottomrule
			\end{tabular}
		}
		\label{tab:epoch}
	\end{table}
	\begin{figure}[t]
		\centering
		\includegraphics[width=0.48\textwidth]{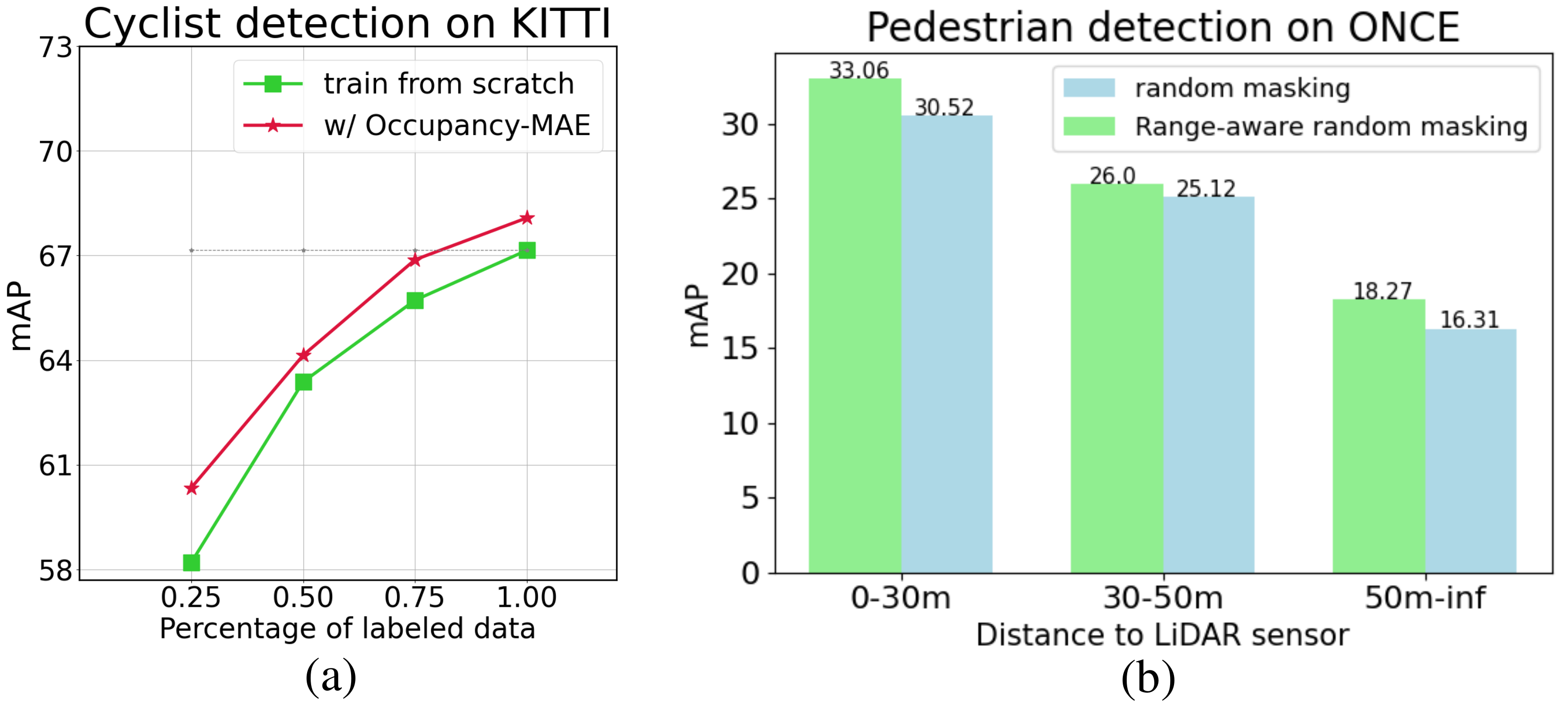}
		\caption{(a) Data efficiency of Occupancy-MAE. (b) Comparison of different masking strategies.}
		\label{fig:percentage2}
	\end{figure}
	
	\begin{table}[t]
		\centering
		\caption{Impacts of the amount of pre-training data.}
		\setlength{\tabcolsep}{2.3mm}
		{
			\begin{tabular}{c|c|c|c|c|c}
				\toprule
				\textbf{Data size} &\textbf{120k} &\textbf{240k} &\textbf{360k} &\textbf{480k}&\textbf{600k}\\
				\midrule
				mAP$\uparrow$ &52.51&52.71& 53.23& 53.57& 53.61  \\
				\bottomrule
			\end{tabular}
		}
		\label{tab:pre-training}
	\end{table}
	
	\begin{table}[t]
		\centering
		\caption{Impacts of masking ratio.}
		\setlength{\tabcolsep}{0.6mm}
		{
			\begin{tabular}{ccc|cccc}
				\toprule
				\multicolumn{3}{c|}{\textbf{Masking ratio}} &\multicolumn{4}{c}{\textbf{mAP}$\uparrow$} \\
				0-30m&30-50m&50m-inf&overall&0-30m&30-50m&50m-inf\\
				\midrule
				0.5 &0.5&0.5&24.07&26.42&22.33&17.57  \\
				\midrule
				0.7 &0.5&0.3&26.29&29.10&23.38&17.80  \\
				0.7 &0.7&0.7&26.57&29.31&24.93&16.01 \\
				\midrule
				0.9 &0.7&0.5&{\bf29.15}&{\bf33.06}&{\bf26.00}&{\bf18.27} \\	
				0.9 &0.7&0.7&27.94&31.25&24.49&18.06 \\	
				0.9 &0.9&0.9&27.49&30.54&25.28&16.11 \\
				\midrule
				0.95 &0.95&0.95&23.68&26.10&21.65&15.98  \\
				\midrule
				0.98 &0.98&0.98&21.61&23.88&19.16&15.74  \\
				
				\bottomrule
			\end{tabular}
		}
		\label{tab:ratio}
	\end{table}
	
	For self-supervised learning on the 3D semantic segmentation task, we designed the lightweight decoder of Occupancy-MAE with only two 3D Deconv layers. Table~\ref{tab:nuscenes_seg} shows that Occupancy-MAE outperforms training from scratch by around 2\% mIoU on nuScenes \emph{val} set. This implies that Occupancy-MAE is well suited for 3D semantic segmentation task since our masked voxel classification objective promotes the model to capture the occupancy distribution of the 3D space, which is also very important for the dense prediction task, i.e. 3D semantic segmentation. Results in Table~\ref{tab:nuscenes_seg} were obtained by retraining Cylinder3D~\cite{cylinder3d}. 
	
	\subsubsection{Multi-object Tracking}
	
	We also validated the performance of the proposed pretraining algorithm on LiDAR-based multi-object tracking tasks. Using GNN-PMB~\cite{gnn-pmb} as the baseline, as shown in Table~\ref{tab:mot}, after employing our proposed Occupancy-MAE pre-training algorithm, the multi-object tracking algorithms with detectors PointPillars~\cite{pointpillars} and CenterPoint~\cite{centerpoint} showed an improvement of approximately 1\% in terms of AMOTA and AMOTP. This indicates that our pre-training algorithm is capable of learning representative features, enhancing 3D object detection performance, and consequently improving multi-object tracking performance.
	
	\subsubsection{Unsupervised Domain Adaptation}
	
	The domain shifts in LiDAR-based 3D perception models are apparent as the LiDARs have different patterns. Evaluation of datasets captured in different locations or sensors results in a drop in model performance. To further investigate the generalization ability of Occupancy-MAE in representation learning, we conduct experiments in two scenarios of unsupervised domain adaptative (UDA) 3D object detection: different collection locations and time (i.e. Waymo $\rightarrow$ KITTI) and different LiDAR ring numbers (i.e. nuScenes $\rightarrow$ KITTI). We compare our method with Oracle, Source-only, and ST3D~\cite{st3d}. We follow the setting of the SOTA UDA method ST3D. We first train Occupancy-MAE on the source and target data simultaneously and then apply the pre-trained model to warm up the object detector in ST3D. Table~\ref{tab:transfer} shows that pre-training with Occupancy-MAE further narrows the gap between UDA and Oracle by about 0.3\% $\sim$ 1\% mAP. Our pre-trained network is voxel-awareness of the object shape of the target data, enabling the UDA model to generate high-quality pseudo-labels for self-training.
	
	\section{Further Analysis}
	We will delve into a comprehensive analysis of the key components of the proposed Occupancy-MAE.
	
	\begin{figure*}[t]
		\centering
		\includegraphics[width=1.0\textwidth]{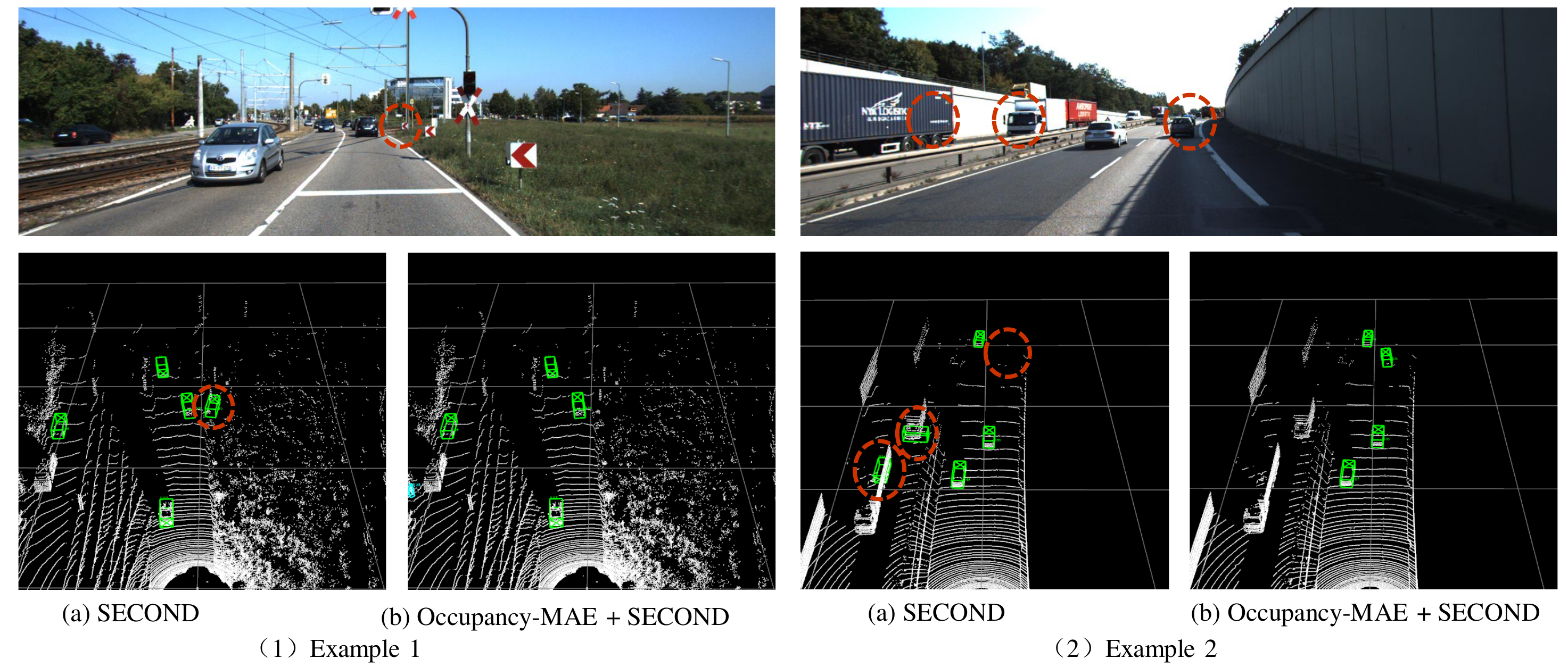}
		\caption{Qualitative results achieved on the KITTI \emph{test} set. With the pre-training of our Occupancy-MAE, the 3D detector can learn more robust features to reduce missed and false detection.}
		\label{fig:visualization}
	\end{figure*}
	\subsubsection{Data-efficient Learner}
	Pre-training enables fine-tuning of models using limited labeled data. To examine the data efficiency of Occupancy-MAE, we conducted experiments with varying amounts of labelled data used for fine-tuning, using SECOND~\cite{second} as the backbone, and evaluated its detection performance on the KITTI \emph{val} set. Our results, shown in Figure~\ref{fig:percentage} and Figure~\ref{fig:percentage2} (a), indicate that using Occupancy-MAE with 50\%, 75\%, and 75\% of labelled data, respectively, results in the same performance as training from scratch with the full dataset for the car, pedestrian, and cyclist classes. With only 25\% of the samples for fine-tuning, our Occupancy-MAE model improves performance by 1\% $\sim$ 2\% mAP over training from scratch, highlighting its data efficiency and ability to reduce the need for costly human-annotated 3D data.
	
	\subsubsection{Range-aware Random Masking}
	We study the effectiveness of the proposed range-aware random masking strategy. For range-aware random masking, the masking ratio for voxels within 0-30 meters, 30-50 meters, and $>$ 50 meters is set to 90\%, 70\%, and 50\%, respectively. We compare it with random masking (the masking ratio is set to 90\% for all voxels). We can see from Figure~\ref{fig:percentage2} (b) that range-aware random masking surpasses random masking by 2.5\%, 0.7\%, and 2.1\% mAP with the ONCE raw set as the pre-training datasets. This demonstrates that the masking ratio for large-scale point clouds should be inversely proportional to the distance from the LiDAR sensor.
	\subsubsection{Occupancy prediction pretext task}
	
	
	The key to successful self-supervised learning lies in selecting a suitable pretext task. In this section, we evaluate the effectiveness of the 3D occupancy prediction pretext task and compare it with the voxel feature regression loss on the KITTI \emph{val} set, which replaces the binary occupancy classification loss.
	The results presented in Table~\ref{tab:task} demonstrate that the performance of the voxel regression task is comparable to training from scratch. However, the classification task outperforms the regression task by 0.2\%, 1.6\%, and 1.4\% in terms of mAP on the car, pedestrian, and cyclist classes, respectively. We believe that the occupancy prediction task is more challenging, and determining whether a voxel contains points is crucial for 3D perception models. By employing the simple 3D occupancy classification task, the pre-trained network becomes more aware of the object's shape, leading to enhanced performance in downstream tasks.
	
	\section{Ablation Studies}
	In this section, we perform more ablation experiments to investigate the individual components of our Occupancy-MAE. All experiments are conducted with the 3D object detector SECOND [45].
	
	\paragraph{Pre-training Times}
	
	In Table~\ref{tab:epoch}, we first study the influence of pre-training times on the KITTI \emph{val} set. The results show that the performance initially increases with increasing pre-training times and then reaches a stable state. In our experiments, we pre-train Occupancy-MAE for three epochs.
	
	\paragraph{Pre-training Data}
	
	We then investigate the impact of the amount of pre-training data. As shown in Table~\ref{tab:pre-training}, the performance of Occupancy-MAE improves with an increase in the amount of unlabeled data on the ONCE \emph{medium} raw set. As autonomous driving companies have access to a large amount of unlabeled data, Occupancy-MAE can utilize this readily available data to enhance the performance of 3D perception in practice.
	
	\paragraph{Masking Ratio}
	In this section, we investigate the influence of the voxel masking ratio on the cyclist class of ONCE \emph{small} raw set. As shown in Table~\ref{tab:ratio}, when increasing the masking ratio to 90\%, the overall performance boosts from 23.27\% to 29.15\%. For the points within 30-50 meters and $>$ 50 meters, the masking ratio of 70\% and 50\% are suitable for downstream tasks, respectively. The proper masking ratio enforces networks to capture the gestalt of the 3D scene to learn representative features. 
	
	\section{Visualization}
	
	Figure~\ref{fig:visualization} shows the visualization of predictions by SECOND~\cite{second} and SECOND with pre-training of our Occupancy-MAE. The limited resolution of LiDAR will lead to missed and false detection of faraway objects. With our Occupancy-MAE, the detectors will learn more robust features to improve detection as shown in Figure~\ref{fig:visualization} (1). Occlusion is another challenging problem in 3D object detection, as shown in Figure~\ref{fig:visualization} (2), the masked voxel classification task pushes Occupancy-MAE to be voxel-aware of the whole shape of objects, thus improving the detection for the Occlusion areas.
	
	\section{Limitations}
	
	Our pre-training target of occupancy prediction relies on dense representations, and 3D convolutions in decoder cannot handle high-resolution feature representations. This will be the focus of our next research phase, where we may consider employing a cascading structure or sparse 3D convolutions for pre-training via masked occupancy prediction.
	
	\section{Conclusion}
	In this paper, we have presented Occupancy-MAE, a self-supervised masked occupancy autoencoding framework designed for pre-training large-scale outdoor LiDAR point clouds. By leveraging vast amounts of unlabeled data in autonomous driving, Occupancy-MAE reduces the reliance on labelled 3D training data. The framework predicts the masked occupancy structure of the entire 3D environment, compelling the network to learn robust features for accurately deducing the masked points. We have also introduced the range-aware random masking strategy, which enhances the pre-training performance for distant objects.
	Extensive experimental results have demonstrated the effectiveness of Occupancy-MAE across various downstream tasks, including 3D object detection, semantic segmentation, multi-object tracking, and unsupervised domain adaptation. Occupancy-MAE has consistently outperformed training from scratch, showcasing its ability to improve the accuracy and performance of perception models in autonomous driving.
	Looking ahead, our future work will explore the pre-training of high-resolution feature maps, 4D scene LiDAR point clouds by incorporating temporal multi-frame data fusion, and exploring to learn general representations across different large-scale outdoor LiDAR point clouds datasets. 
	Overall, Occupancy-MAE represents a promising approach to address the challenges of labelled data scarcity in autonomous driving, offering a cost-effective and efficient solution for enhancing the 3D perception ability of autonomous vehicles.

	
	\bibliographystyle{IEEEtran}
	\bibliography{egbib}

\end{document}